\documentclass{bmvc2k}

\usepackage{url}
\usepackage{subcaption}
\usepackage[draft]{pgf}
\usepackage{latexsym}
\usepackage{amsmath}
\usepackage{multirow}
\usepackage{url}

\usepackage{graphicx}
\usepackage{latexsym}
\usepackage{booktabs}
\usepackage{url}
\usepackage{adjustbox}
\usepackage{color}
\usepackage{comment}
\usepackage{array}
\newcolumntype{L}[1]{>{\raggedright\arraybackslash}p{#1}}
\usepackage[colorinlistoftodos]{todonotes}
\usepackage{rotating}
\usepackage{enumerate}

\title{End-to-end Image Captioning Exploits Multimodal Distributional Similarity}

\addauthor{Pranava Madhyastha}{p.madhyastha@sheffield.ac.uk}{1}
\addauthor{Josiah Wang}{j.k.wang@sheffield.ac.uk}{2}
\addauthor{Lucia Specia}{l.specia@sheffield.ac.uk}{3}

\addinstitution{
 Department of Computer Science\\
 The University of Sheffield\\
 Sheffield, UK
}

\runninghead{~Madhyastha \etal}{Image Captioning Exploits Distributional Similarity~}

\def\etal{\emph{et al}\bmvaOneDot}

\begin{document}

\maketitle
\begin{abstract}
We hypothesize that end-to-end neural image captioning systems work seemingly well because they exploit and learn `distributional similarity' in a multimodal feature space by mapping a test image to similar training images in this space and generating a caption from the same space. To validate our hypothesis, we focus on the `image' side of image captioning, and vary the input image representation but keep the RNN text generation component of a CNN-RNN model constant. Our analysis indicates that image captioning models (i) are capable of separating structure from noisy input representations; (ii) suffer virtually no significant performance loss when a high dimensional representation is compressed to a lower dimensional space; (iii) cluster images with similar visual and linguistic information together. Our findings indicate that our distributional similarity hypothesis holds. We conclude that regardless of the image representation used image captioning systems seem to match images and generate captions in a learned joint image-text semantic subspace. 
\end{abstract}

\section{Introduction}

Image description generation, or image captioning (IC), is the task of automatically generating a textual description for a given image. The generated text is expected to describe, in a single sentence, what is visually depicted in the image, for example the entities/objects present in the image, their attributes, the actions/activities performed, entity/object interactions (including quantification), the location/scene, etc. (e.g. ``\textit{a man riding a bike on the street}''). Significant progress has been made with \emph{end-to-end} approaches to tackling this problem, where parallel image--description datasets such as Flickr30k~\citep{young2014image} and  MSCOCO~\citep{mscoco2015} are used to train a CNN-RNN based neural network IC system~\citep{vinyals2016show,karpathy2015deep,xu2015show}. Such systems have demonstrated impressive performance in the COCO captioning challenge\footnote{\url{http://cocodataset.org/\#captions-challenge2015}} according to automatic metrics, seemingly even surpassing human performance in many instances (e.g.\ CIDEr score $>1.0$ %
vs. human's 0.85)~\citep{mscoco2015}. However, in reality, the performance of end-to-end systems is still far from satisfactory according to metrics based on human judgement\footnote{\url{http://cocodataset.org/\#captions-leaderboard}}. Thus, despite the progress, this task is currently far from being a solved problem.

In this paper, we challenge the common assumption that end-to-end IC systems are able to achieve strong performance because they have learned to `understand' and infer semantic information from visual representations, i.e.\ they can for example deduce that ``\textit{a boy is playing football}" purely by learning directly from mid-level image features and the corresponding textual descriptions in an implicit manner, without explicitly modeling the presence of  {\em boy}, {\em ball}, {\em green field}, etc. in the image. It is believed that the IC system has managed to infer that the phrase \textit{green field} is associated  with some `green-like' area in the image and is thus generated in the output description, or that the word \textit{boy} is generated because of some CNN activations corresponding to a young person. However, there seems to be no concrete evidence that this is the case. Instead, we hypothesize that the apparently strong performance of end-to-end systems is attributed to the fact that they exploit the \emph{distributional similarity} property in a multimodal feature space.
To our best knowledge, our paper gives the first empirical analysis on visual representations for the task of image captioning.

What we mean by `distributional similarity' is that IC systems essentially attempt to find images from the training set that are most similar to a test image, and generate a caption from the most similar training instances (or generate a `novel' description from a combination of training instances, for example by `averaging' the descriptions). Previous work has alluded to this observation ~\citep{kaparthy2016connecting,vinyals2016show}, but it has not been thoroughly studied. This phenomenon could also be in part attributed to the fact that the datasets are repetitive and simplistic, with an almost constant and predictable linguistic structure~\citep{lebret2015phrase,devlin2015language,vinyals2016show}. Thus, while IC systems perform very well at the task of matching images to captions at surface-level, they do not truly understand images or language and use this understanding to generate image descriptions. Such misconception can deter true progress in the field. This paper aims to draw attention to this issue and to the importance of understanding how IC systems work and with that support work towards progress in the field that goes beyond optimizing for metrics to achieve state-of-the-art performance.

It is worth noting that we are interested in demonstrating the phenomenon of distributional similarity in IC, rather than achieving or improving state-of-the-art performance. As such, we do not resort to fine-tuning or extensive hyperparameter optimization or ensembles. Therefore, our model is not comparable to state-of-the-art models such as~\citet{vinyals2016show}, which optimize IC by fine-tuning the image representations, exploring beam size, scheduled sampling, and using ensemble models. Instead, we vary only the image representation to demonstrate that end-to-end IC systems utilize distributional similarity on the image side to generate captions, regardless of the image representation used. %

Our main contributions are: %
\begin{enumerate}[(a)]
\item \textbf{An IC experiment}
 where we vary the input image representation but keep the RNN text generation model component constant (Section~\ref{sec:ic}). This experiment demonstrates that regardless of the image representation (a continuous image embedding or a sparse, low-dimensional vector), end-to-end IC systems seem to utilize a visual-semantic subspace for IC.
        \item The introduction of \textbf{pseudo-random vectors} derived from object-level
        representations as a means to evaluate IC systems. Our results show
        that end-to-end models in this framework are remarkably capable of
        separating structure from noisy input representations.
        \item An experiment where IC models are conditioned on image
        representations~\textbf{factorized and compressed to a lower
        dimensional space} (Section~\ref{sec:factorization}). We show that high        dimensional image embeddings that are factorized to a lower dimensional
        representation and used as input to an IC model result in virtually no
        significant loss in performance, further strengthening our claim that
        IC models perform similarity matching rather than image understanding.
        \item An \textbf{analysis of different image representations and their
        transformed representations} (Section~\ref{sec:transformed}). We visualize the initial visual subspace and the learned joint visual semantic subspace and observe that the visual semantic subspace has learned to cluster images with similar visual and linguistic information together, further validating our claims of distributional similarity. 
        \item An experiment where the IC model is \textbf{tested on an out-of-domain dataset} (Section~\ref{sec:domaindependency}), which has a slightly different image distribution. We observe that models
        show better performance on test sets that have a similar distribution as the training. Their performance deteriorates when the distributions are even slightly different.
\end{enumerate}

Overall, our study demonstrates that end-to-end IC models implicitly learn and exploit multimodal similarity spaces rather than performing actual image understanding.

\section{Model setting}
\label{sec:modelsetting}

For the experiments in Section~\ref{sec:ic},
we base our implementation on the end-to-end approach by~\citet{karpathy2015deep}. 
We use the LSTM~\citep{hochreiter1997long} based language model as described in~\citet{zaremba2014recurrent}, which is conditioned on the image information. For that, we first perform a linear projection of the image representation followed by a non-linearity: 
\begin{equation}
Im_{feat} = \sigma{(W{\cdot}I_m)}
\label{affine}
\end{equation}

Here, $I_m \in \mathcal{R}^d$ is the $d$-dimensional initial image representation, $W \in \mathcal{R}^{n{\times}d}$ is the linear transformation matrix, $\sigma$ is the non-linearity. We use Exponential Linear Units~\citep{clevert2015fast} as the non-linear activation in all our experiments. Following~\citet{vinyals2015show}, we initialize the LSTM based caption generator with the projected image feature.

\paragraph{Training and Inference}

The image caption generator is trained to generate sentences conditioned on the image representation by minimizing a cross-entropy loss, i.e., the sentence-level loss corresponds to the sum of the negative log likelihood of the correct word being generated at each time step: 

\begin{equation}
\Pr(S{\mid}Im_{feat};\theta) = \sum_t{\log(\Pr(w_t|w_{t-1}..w_0;Im_{feat}))}
\label{likelihood}
\end{equation}

\noindent{where $\Pr{(S{\mid}Im_{feat};\theta)}$ is the sentence-level loss conditioned on the image feature $Im_{feat}$ and $\Pr(w_t)$ is the probability of the word at time step $t$. 
This is trained with standard teacher forcing as described in \citet{sutskever2014sequence} where the correct word information is fed to the next state in the LSTM. 

Inference is typically performed with approximation techniques like beam search or sampling~\citep{karpathy2015deep,vinyals2015show}. In this paper, as we are mainly interested in studying the effect of different image representations, we focus on the language output that the models can most confidently produce. Therefore, unless stated otherwise we generate captions using a greedy ${\arg}{\max}$ approach.

\section{Image captioning with different image representations}
\label{sec:ic}

In this section, we verify our hypothesis that a `distributional similarity' space exists in end-to-end IC systems. Such systems attempt to match image representations in order to condition the RNN decoder to generate captions that are similar to the closest images, rather than actually understanding the image in order to describe it. We keep the IC model constant (Section~\ref{sec:modelsetting}) across experiments and vary only the image representation used. The different representations we experimented with are described in what follows.

\subsection{Lower-bound image representation}
\label{sec:representations}

\paragraph{Random:} We condition the LSTM on a 300-dimensional vector comprising random values sampled uniformly between $[0,1)$\footnote{We also tried using 1,000-dimensions, which yielded similar but slightly poorer results.}. This feature essentially gives us a worst-case image feature and thus provides an artificial lower bound.

\subsection{Representations from image-level classification}

We compare two CNNs -- \emph{VGG19}~\citep{simonyan2015very} and \emph{ResNet152}~\citep{he2016deep} -- both pre-trained on the ILSVRC challenge data~\citep{russakovsky2015imagenet}. We explore various representations derived from these CNNs:

\paragraph{Penultimate layer (\emph{Penultimate}):}
Most previous attempts to IC use the output of the penultimate layer of a CNN pre-trained on ILSVRC.  
Previous work motivates using `off-the-shelf' feature extractors 
in the framework of transfer learning \citep{razavian2014cnn,donahue2014decaf}. Such features have often been applied to image captioning~\citep{mao2014deep,karpathy2015deep,xu2015show,gao2015you,vinyals2015show,donahue2015long} and have been shown to produce state-of-the-art results. Therefore, we extract the \textbf{\emph{fc7}} layer from \emph{VGG19} ($4,096D$) and the \textbf{\emph{pool5}} layer from \emph{ResNet152} ($2,048D$) for each image. 

\paragraph{Class prediction vector (\emph{Softmax}):}
We also investigate higher-level image representations where each element in the vector is the estimated posterior probability of an object category appearing in that image. Note that the categories may not directly correspond to the captions in the dataset. 
While there are alternative methods that fine-tune the image network on a new set of object classes extracted in ways that are directly relevant to the captions~\citep{fang2015captions,wu2016value}, we study the impact of off-the-shelf prediction vectors on the IC task.
The intuition is that category predictions from pre-trained CNN classifiers may also be beneficial for IC, alongside the standard approach of using mid-level features from the penultimate layer. Therefore, for each image, we use the predicted category posterior distributions of \emph{VGG19} and \emph{ResNet152} for $1,000$ object categories.

\paragraph{Object class word embeddings (\emph{Top-$k$}):} Here we experiment with a method that utilizes the averaged word representations of top-$k$ predicted object classes. We first obtain \emph{Softmax} predictions using \emph{ResNet152} for $1,000$ object categories (synsets) per image.
We then select the objects that have a posterior probability score $>5\%$ and use the $300$-dimensional pre-trained word2vec~\citep{mikolov2013distributed} representations\footnote{\url{https://code.google.com/archive/p/word2vec/}} to obtain the averaged vector over all retained object categories. This is motivated by the observation that averaged  word embeddings can represent semantic-level properties and are useful for classification tasks~\citep{arora2016simple}.

\subsection{Representations from object-level detections}

We also explore representing images using information from object \emph{detectors} that identify \emph{instances} of object categories present in an image, rather than a global, image-level classification. This can potentially provide for a richer and more informative image representation. For this we use:
\begin{itemize}
\item \emph{ground truth} (\textbf{\emph{Gold}}) region annotations for instances of 80 pre-defined categories provided with MSCOCO. It is worth noting that these were annotated independently of the image captions, i.e. people writing the captions had no knowledge of the 80 categories. As such, there is no direct correspondence between the region annotations and image captions.
\item a state-to-the-art object detector \emph{YOLO}~\citep{redmon2016yolo9000}, pre-trained on MSCOCO for 80 categories (\textbf{\emph{YOLO-Coco}}), and on MSCOCO and ILSVRC for over 9,000 categories (\textbf{\emph{YOLO-9k}}). We use \emph{YOLOv2}.
\end{itemize}

We explore several representations derived from instance-level object class annotations or detectors above:

\paragraph{Bag of objects (\emph{BOO}):} 

We represent each image as a sparse `bag of objects' vector, where each element represents the frequency of occurrence for each object category in the image (\textbf{\emph{Counts}}). We also explore an alternative representation where we only encode the presence or absence of the object category regardless of its frequency (\textbf{\emph{Binary}}) to determine whether it is important to encode object counts in the image. 
These representations help us examine the importance of explicit object categories and in a sense interactions between object categories (e.g. \textit{dog} and \textit{ball}) in the image representation. We investigate whether such a sparse and high-level \emph{BOO} representation is actually sufficient for generating image captions. It is also worth noting that \emph{BOO} is different from the \emph{Softmax} representation above as it encodes the \emph{number} of object occurrences, not the \emph{confidence} of class predictions at image level. We compare \emph{BOO} representations derived from the \textbf{\emph{Gold}} annotations (\textbf{\emph{Gold-Binary}} and \textbf{\emph{Gold-Counts}}) and both \textbf{\emph{YOLO-Coco}} and \textbf{\emph{YOLO-9k}} detectors (\textbf{\emph{Counts}} only). 

\paragraph{Pseudo-random vectors:} To further probe the capacity of the model to discern image representations in an image distributional similarity space, we propose a novel experiment in which we examine a type of representation where \emph{similar images are represented using similar random vectors}, which we term as \emph{pseudo-random vectors}. We form this representation from \textbf{\emph{BOO Gold-Counts}} and \textbf{\emph{BOO Gold-Binary}}. More specifically, $Im_{feat} = \sum_{o \in \text{Objects}} f \times \phi_{o}$, where $\phi_{o} \in \mathcal{R}^d$ is an object-specific random vector and $f$ is a scalar representing counts of the object category. In the case of \textbf{\emph{Pseudorandom-Counts}}, $f$ is the frequency counts from \textbf{\emph{Gold-Counts}}. In the case of \textbf{\emph{Pseudorandom-Binary}}, $f$ is either $0$ or $1$ based on \textbf{\emph{Gold-Binary}}. We use $d = 120$ for these experiments. Intuitively, these pseudo-random vectors appear random and noisy in the representational space as a result of the composition of (random) object category vectors, more specifically the multiplication of object category vectors by their frequency of occurrence and the addition of vectors across multiple object categories. We use these vectors to demonstrate that end-to-end IC models are capable of separating structure from noise, and thus exploit the distributional similarity property in a multimodal feature space.

\subsection{Datasets and experimental setup}
\label{sec:setup}

\paragraph{Dataset}
We evaluate image captioning conditioned on different representations on the most widely used dataset for IC, MSCOCO~\citep{mscoco2015}. The dataset consists of $82,783$ images for training, with at least five captions per image, totaling to $413,915$ captions. %
We perform model selection on a $5000$-image development set and report the results on a $5000$-image test set using standard, publicly available splits\footnote{\url{http://cs.stanford.edu/people/karpathy/deepimagesent}} of the MSCOCO validation dataset as in previous work~\citep{karpathy2015deep}.

\subsection{Image captioning results}
\label{sec:results}

We report results of IC on MSCOCO in Table \ref{tab:classembeddings}, where the IC model (Section~\ref{sec:modelsetting}) is conditioned on the various image representations described in Section~\ref{sec:representations}.  
As expected, using random image embeddings 
clearly does not provide any useful information and performs poorly.
The CNN \emph{softmax} representations with the same set of $1,000$ object classes
(\emph{VGG19} and \emph{ResNet152}) have very
similar performance. We note that the posterior distribution may not directly correspond to words in the captions, i.e. many words and concepts  are not contained in the set of object classes. Our results differ from those by \citet{wu2016value} and~\citet{yao2016boosting} where the object classes have been fine-tuned to correspond directly to the caption vocabulary. %

\begin{table}[htbp]
\begin{center}
\resizebox{0.7\linewidth}{!}{%
\begin{tabular}{*{9}{c}}
\toprule
& \textbf{Representation} & \textbf{B-1} & \textbf{B-2} & \textbf{B-3} & \textbf{B-4} & \textbf{M} & \textbf{C} & \textbf{S}\\
\midrule
& Random & 0.48 & 0.24 & 0.11 & 0.07 & 0.11 & 0.07 & 0.03 \\ %
\midrule
\multirow{2}{*}{\footnotesize{Softmax}} & VGG19 & 0.62 & 0.43 & 0.29 & 0.19 & 0.20 & 0.61 & 0.13 \\
& ResNet152 & 0.62 & 0.43 & 0.29 & 0.19 & 0.20 & 0.62 & 0.12 \\ %
\midrule
    \multirow{2}{*}{\footnotesize{Penultimate}} & VGG19 (fc7) & 0.65 & 0.46 & 0.32 & 0.22 & 0.21 & 0.69 & 0.14\\
& ResNet152 (pool5) & 0.66 & 0.48 & 0.33 & 0.23 & 0.22 & 0.74 & 0.15\\ %
\midrule
\footnotesize{Embeddings} & Top-$k$ & 0.62 & 0.42 & 0.28 & 0.19 & 0.20 & 0.63 & 0.13 \\
\midrule
\multirow{4}{*}{\footnotesize{BOO}}& Gold-Binary & 0.65 & 0.47 & 0.32 & 0.22 & 0.22 & 0.75 & 0.15 \\ %
& Gold-Counts & 0.67 & 0.48 & 0.33 & 0.23 & 0.22 & 0.81 & 0.16\\
& YOLO-Coco & 0.65 & 0.46 & 0.32 & 0.22 & 0.22 & 0.75 & 0.15\\
& YOLO-9k & 0.64 & 0.45 & 0.31 & 0.21 & 0.20 & 0.68 & 0.13\\
\midrule
\multirow{2}{*}{\footnotesize{Pseudo-random}} & Pseudorandom-Binary & 0.65 & 0.46 & 0.31 & 0.21 & 0.21 & 0.73 & 0.14 \\ %
 & Pseudorandom-Counts & 0.67 & 0.48 & 0.34 & 0.23 & 0.22 & 0.80 & 0.15 \\ %
\bottomrule
\end{tabular}
}
\end{center}
\caption{Results on the MSCOCO test split, where we vary only the image representation and keep other parameters constant. The captions are generated with $beam=1$. We report \textbf{B}LEU (1-4), \textbf{M}eteor, \textbf{C}IDEr and \textbf{S}PICE scores.}
\label{tab:classembeddings}
\end{table}

The performance of the~\emph{pool5} image representations shows a similar trend for \emph{VGG19} and \emph{ResNet152}, with \emph{ResNet152} achieving slightly better scores than \emph{VGG19}. We posit that the representations from the image network trained on object classes are able to capture more fine-grained image details. 

The performance of the averaged top-$k$ word embeddings is similar to that of the \emph{Softmax} representation. This is interesting, since the averaged word representational information is mostly noisy: we combine top-$k$ synset-level information into one single vector; however, it still performs competitively.

The performance of the \emph{BOO} sparse $80$-dimensional annotation vector is better than all other image representations judging by the CIDEr score. We note again that this occurs despite the fact that the annotations may not directly correspond to the semantic information in the captions or the images. The sparse representational information is indicative of the presence of only a subset of potentially useful objects. We notice two distinct patterns, a marked difference with \emph{Binary} and \emph{Count} representations. This takes us back to the motivation that image captioning  requires information about objects, as well as interactions between objects and their attributes. Although our representation is really sparse on the object interactions, it captures the basic concept of the presence of more than one object of the same kind, and thus provides extra information. A similar trend was observed by \citet{WangEtAl:2018}, who further explored encoding the geometric and size information of objects into the representation, and by \citet{obj2text}, who learn interactions using a specified object-layout RNN. 

We also notice that using predicted objects using \emph{YOLOCoco} performs better than using \emph{YOLO9k}. This is probably expected as \emph{YOLOCoco} was trained on the same dataset hence producing better object proposals. We also observed that \emph{YOLO9k} had a significant number of objects predicted for the test images that had not been seen in the training set (around 20\%).  

The most surprising result is the performance of the pseudo-random vectors. We notice that both the \emph{pseudo-random binary} and  \emph{pseudo-random count} vectors perform almost as well as the \emph{Gold} objects. This suggests that the conditioned RNN is able to remove noise and learn some sort of a common `visual-linguistic' semantic subspace.

\begin{table}
    \begin{minipage}[b]{0.48\hsize}\centering
    \begin{center}
    \resizebox{\linewidth}{!}{
	\begin{tabular}{*{8}{c}}
	\toprule
	\textbf{Method} & \textbf{B-1} & \textbf{B-2} & \textbf{B-3} & \textbf{B-4} & \textbf{M} & \textbf{C} & \textbf{S}\\
	\midrule
	PCA & 0.66 & 0.48 & 0.34 & 0.24 & 0.22 & 0.75 & 0.15 \\ %
	ICA & 0.66 & 0.48 & 0.34 & 0.24 & 0.22 & 0.74 & 0.15 \\ %
	PPCA & 0.66 & 0.48 & 0.34 & 0.24 & 0.22 & 0.76 & 0.15 \\ %
    \midrule
    FULL & 0.66 & 0.48 & 0.33 & 0.23 & 0.22 & 0.74 & 0.15\\
	\bottomrule
	\end{tabular}
    }
    \end{center}
    \caption{Performance of compressed Pool5 representations.}
        \label{tab:factors}
    \end{minipage}
    \hfill
    \begin{minipage}[b]{0.45\hsize}\centering
    \begin{center}
    \resizebox{\linewidth}{!}{
    \begin{tabular}{*{7}{c}}
    \toprule
    \textbf{Model} & \textbf{B-1} & \textbf{B-2} & \textbf{B-3} & \textbf{B-4} & \textbf{M} & \textbf{C}\\
    \midrule
    Pool5 & 0.60 & 0.41 & 0.26 & 0.17 & 0.14 & 0.29 \\ %
    SC & 0.62 & 0.42 & 0.28 & 0.18 & 0.17 & 0.35 \\ %
    TDBU & 0.60 & 0.40 & 0.26 & 0.17 & 0.17 & 0.34 \\ %
    \bottomrule
    \end{tabular}
    }
    \end{center}
    \caption{Performance of models on Flickr30k.}%
    \label{tab:flickr}
    \end{minipage}
\end{table}

\section{Analysis of distributional similarity in IC}
\label{sec:analysis}

In what follows we present further analyses on the different image representations to gain a better understanding of such representations and demonstrate our distributional similarity hypothesis. 

\subsection{Factorizing representations} 
\label{sec:factorization}

In Section~\ref{sec:results} we observed encouraging results from the bag of objects representation despite it being sparse, low-dimensional, and only partially relevant to captions. Interestingly, using pseudo-random vectors derived from a bag of objects also resulted in good performance despite the added noise. This leads to the question: are high-dimensional vectors necessary or relevant? To answer this question, we evaluate whether the performance of the model is significantly poorer if we reduce the dimensionality of the initial 
 high dimensional representation. We experiment with three exploratory factor analysis-based methods -- Principal Component Analysis (PCA)~\citep{pca}, Probabilistic Principal Component Analysis (PPCA)~\citep{ppca} and Independent Component Analysis (ICA)~\citep{ica}. In all cases, we obtain $80$-dimensional factorized representations from \emph{ResNet152 pool5} ($2048D$), which is commonly used in IC.
We summarize our results in Table~\ref{tab:factors}. We observe that the representations obtained by all of the factored models seem to retain the necessary representational power to produce appropriate captions, equivalent to the original representation. This seems contradictory, as we expected a loss in information content when compressing it to arbitrary $80$-dimensions. This experiment indicates that the model is not explicitly utilizing the full expressiveness of the full $2048$-dimensional representations. The model is able to learn from seemingly weak, structured information and can achieve performance that is close to that achieved using the full representation. 

\begin{figure}[th]
  \begin{center}
     \resizebox{\linewidth}{!}{
    \begin{subfigure}[b]{0.51\linewidth}
      \begin{tabular}{c@{}c@{}c}
        \parbox[c]{0.45\linewidth}{\includegraphics[width=\linewidth]{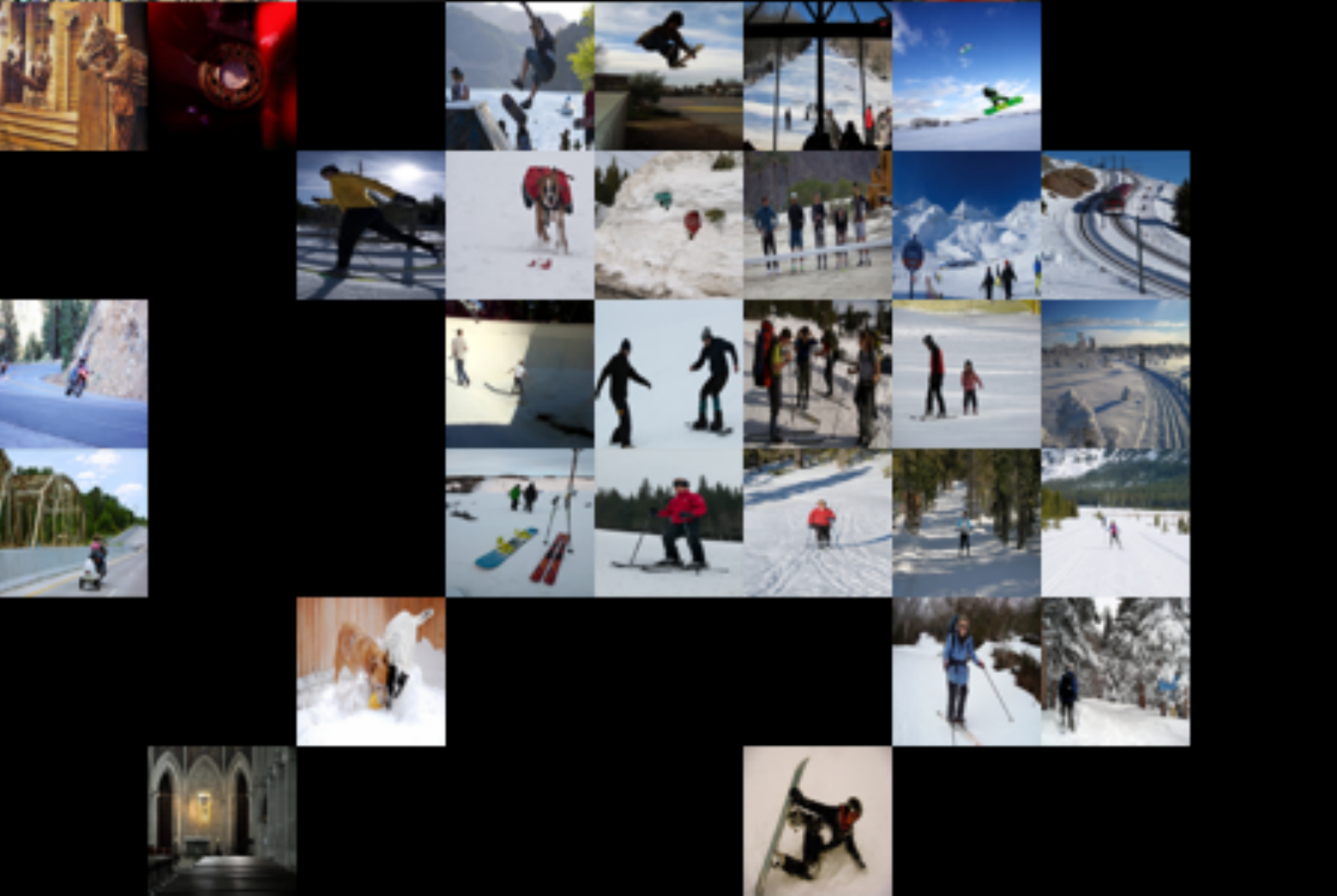}} &
    $\rightarrow$ &
        \parbox[c]{0.45\linewidth}{\includegraphics[width=\linewidth]{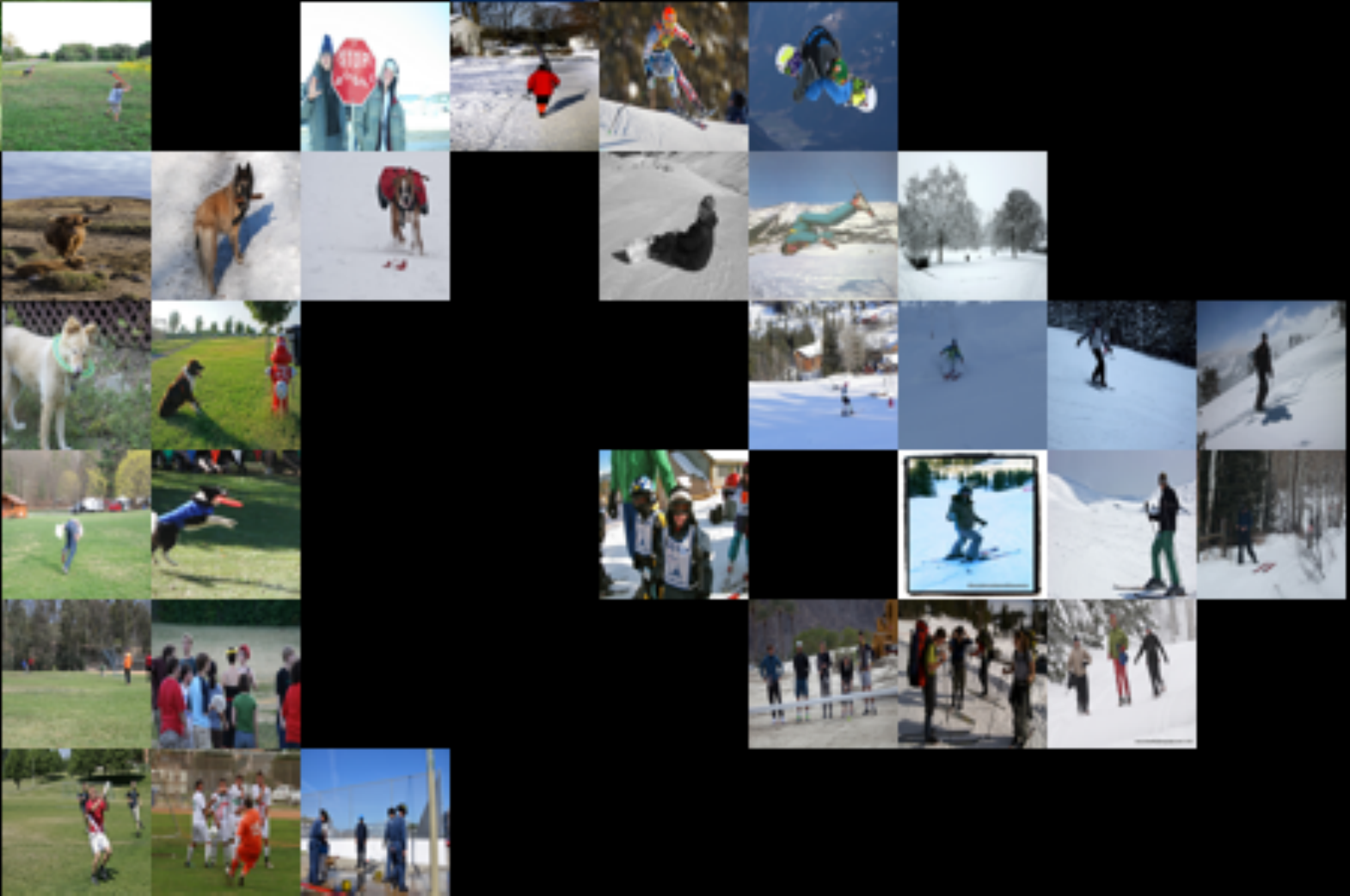}}
      \end{tabular}
      \caption{Pool5\label{fig:tsne_pool5}}
    \end{subfigure}
    \hfill
    \begin{subfigure}[b]{0.48\linewidth}
      \begin{tabular}{c@{}c@{}c}
        \parbox[c]{0.45\linewidth}{\includegraphics[width=\linewidth]{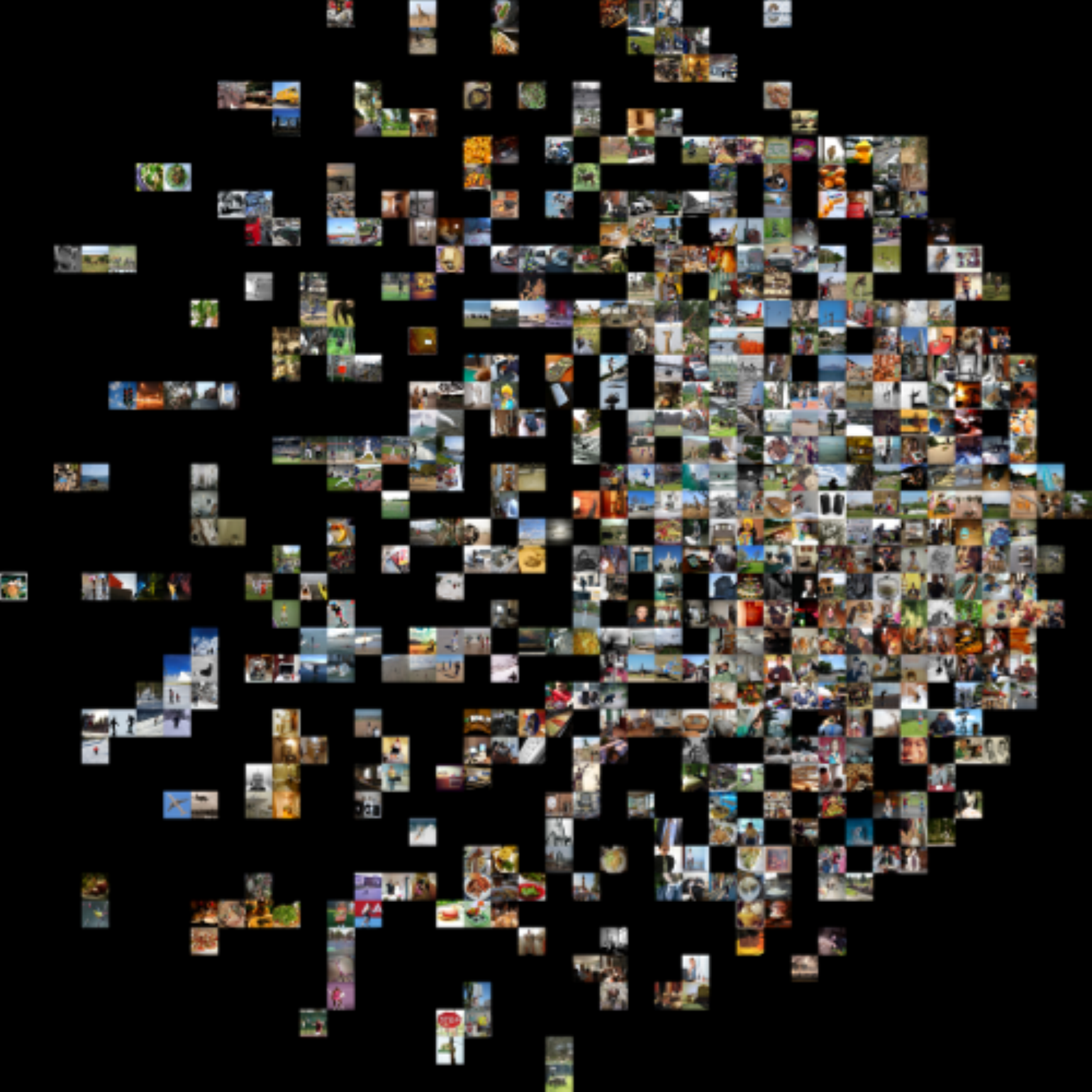}} &
    $\rightarrow$ &
        \parbox[c]{0.45\linewidth}{\includegraphics[width=\linewidth]{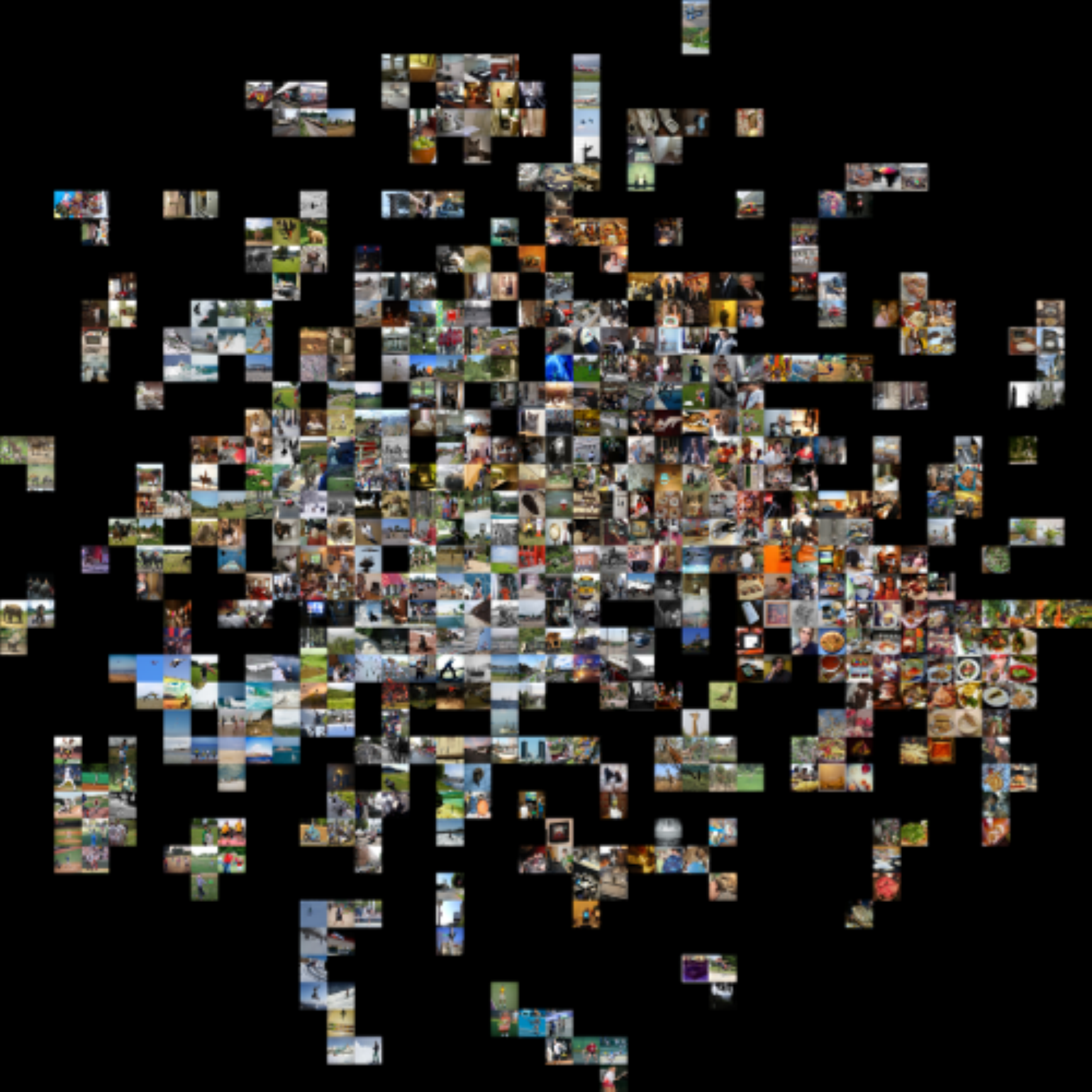}}
      \end{tabular}
      \caption{Softmax\label{fig:tsne_softmax}}   
    \end{subfigure}
    }
    \resizebox{\linewidth}{!}{    
    \begin{subfigure}[b]{0.70\linewidth}
       \begin{tabular}{c@{}c@{}c@{}c@{}c}
        \parbox[c]{0.30\linewidth}{\includegraphics[width=\linewidth]{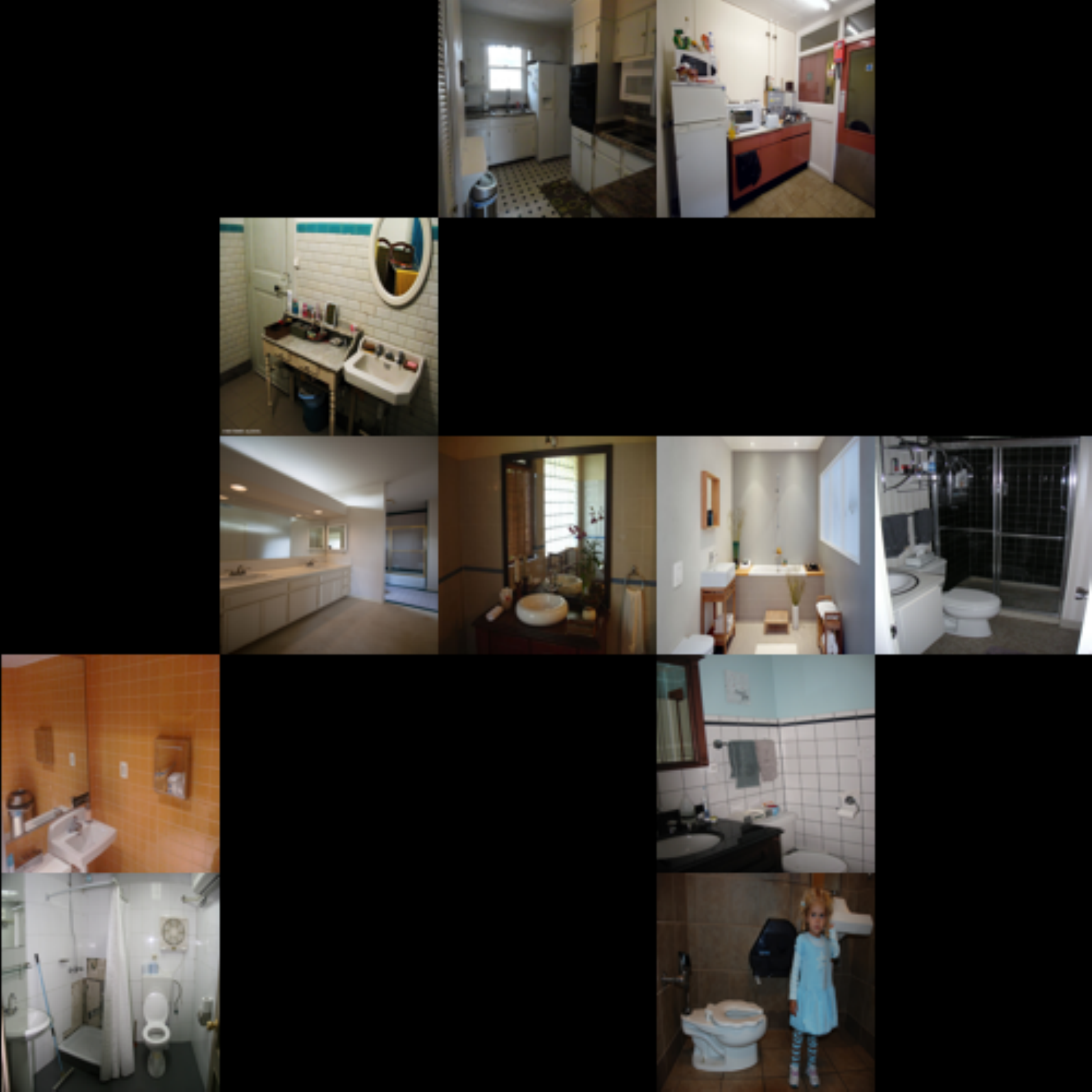}} &
         $\rightarrow$ &
        \parbox[c]{0.30\linewidth}{\includegraphics[width=\linewidth]{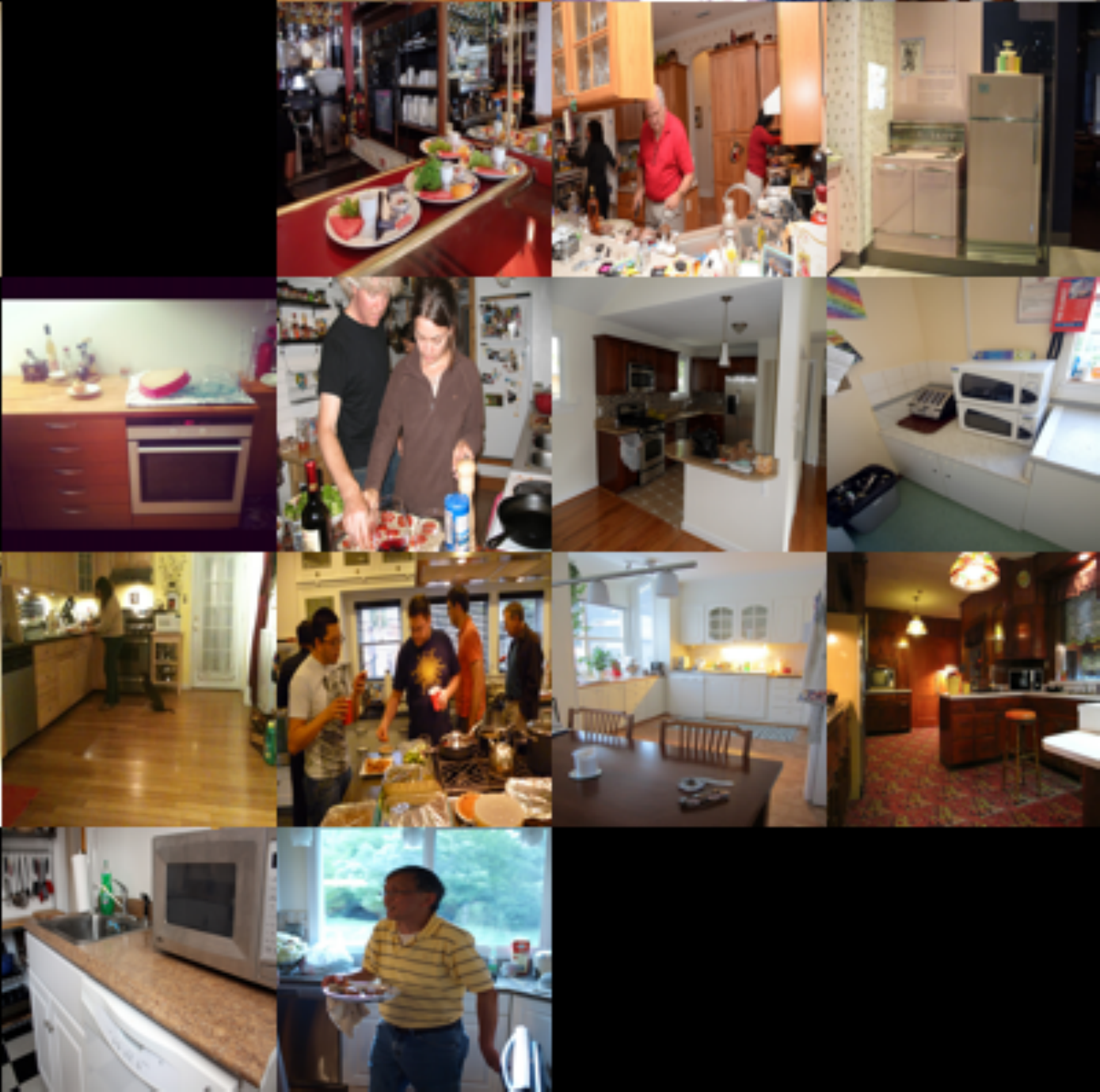}} &  
        ... &
        \parbox[c]{0.30\linewidth}{\includegraphics[width=\linewidth]{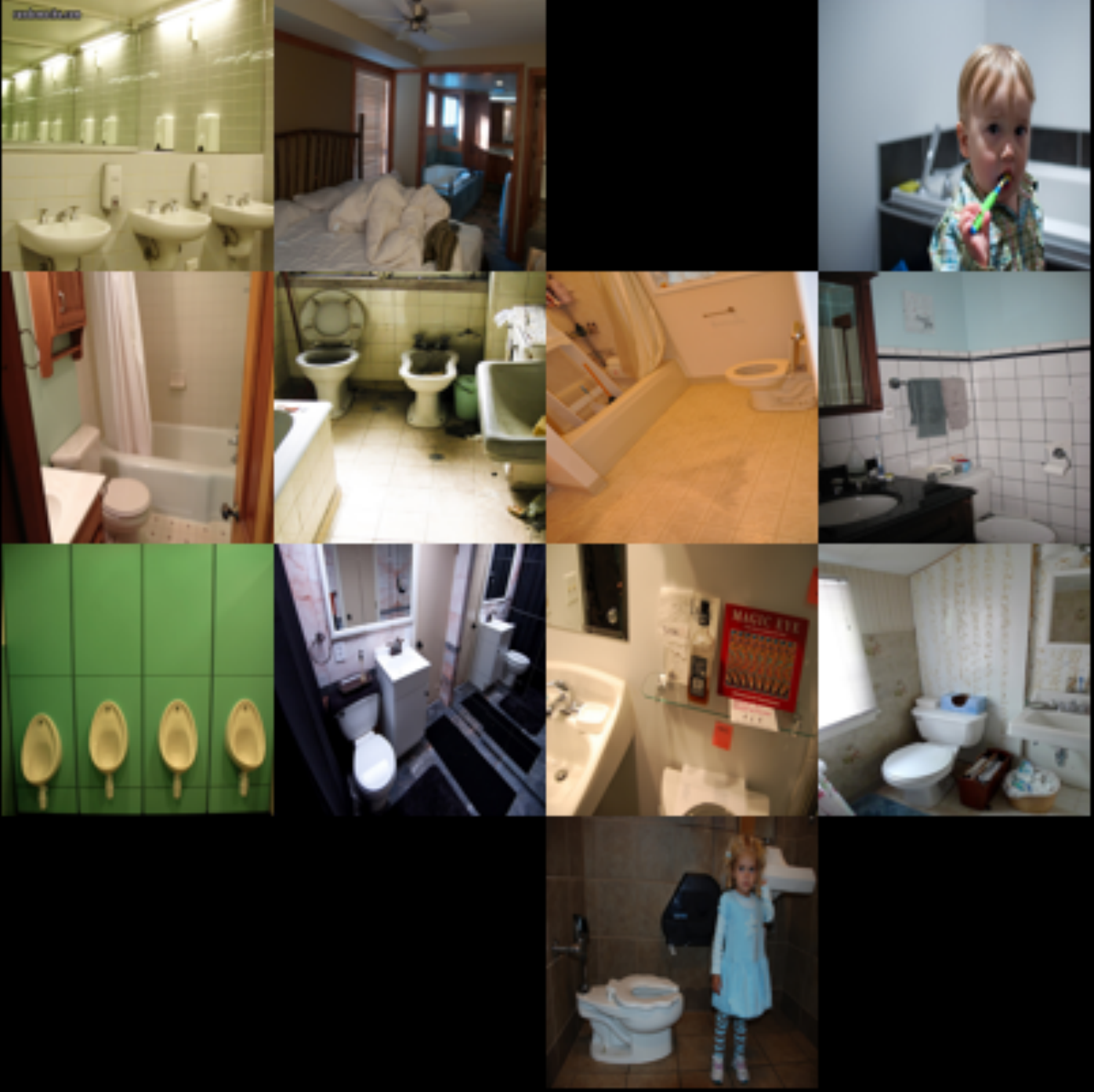}}     
      \end{tabular}
      \caption{Bag of objects\label{fig:tsne_boc}}   
    \end{subfigure}
    \hfill    
    \begin{subfigure}[b]{0.50\linewidth}
       \begin{tabular}{c@{}c@{}c}
        \parbox[c]{0.45\linewidth}{
        \includegraphics[width=\linewidth]{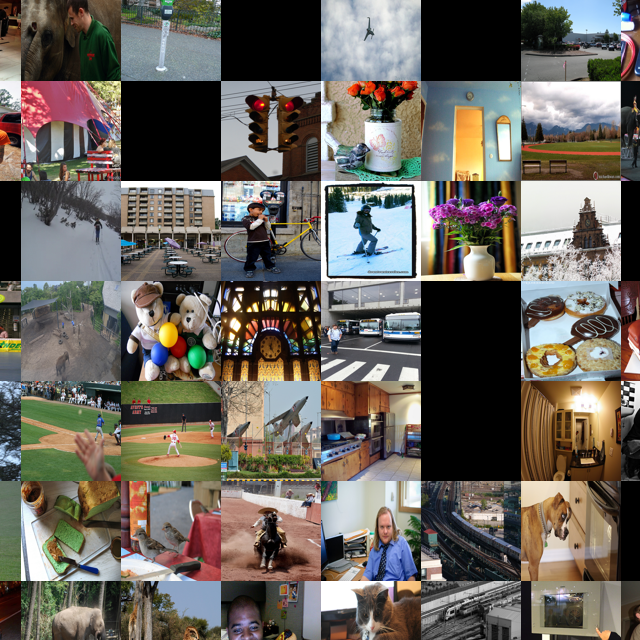}} &
    $\rightarrow$ &
        \parbox[c]{0.45\linewidth}{    
        \includegraphics[width=\linewidth]{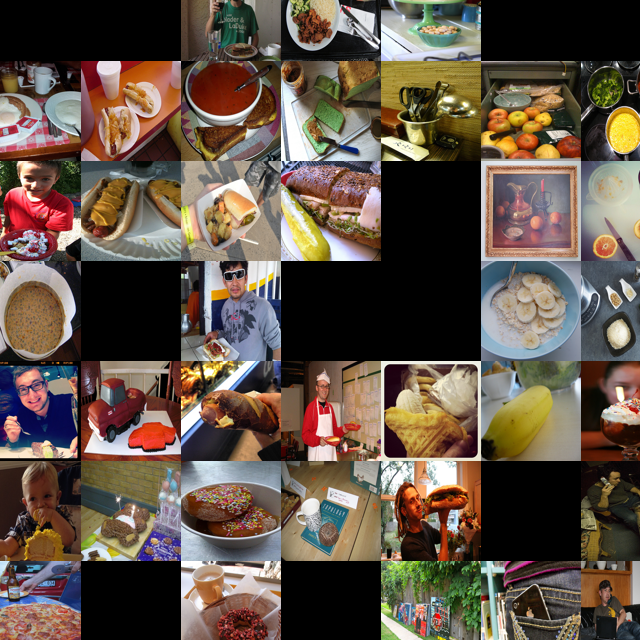}}
      \end{tabular}
      \caption{Pseudo-random\label{fig:tsne_pseudorandom}}
    \end{subfigure}    
    }
  \end{center}
\caption{Visualization of the t-SNE projection of initial representational
    space (left) vs.\ the transformed representational space (right). Please see
 \url{https://github.com/sheffieldnlp/whatIC} for original images.} %
\label{fig:tsne}
\end{figure}

\subsection{Analyzing transformed image representations}
\label{sec:transformed}

Considering our earlier hypothesis as proposed in Section~\ref{sec:results} whereby the conditioned RNN is learning some sort of a common `visual-linguistic' semantic space, we explore the difference in representations in the initial representational space ($I_m$ in Equation~\ref{affine}) and the transformed representational space ($Im_{feat}$ in Equation~\ref{affine}). The transformation matrix $W$ (Equation~\ref{affine})  is learned jointly as a subtask of the image captioning. We posit that image representations in the  $256$-dimensional transformed space will be more semantically coherent with respect to both images and captions. To visualize the two representational spaces, we use \emph{Barnes-Hut t-SNE}~\citep{maaten2008visualizing} to compute a $2$-dimensional embedding over the test split. 
In general, we found that images are initially clustered by visual similarity (\textit{Pool5}) and semantic similarity (\textit{Softmax}, \textit{BOO}). After transformation, we observe that some linguistic information from the captions has produced different types of clusters.
Figure~\ref{fig:tsne} highlights some interesting observations regarding the changes in clustering across three different representations. For \textit{Pool5}, images seem to be clustered by their visual appearance, for example snow scenes in Figure~\ref{fig:tsne_pool5}, regardless of the subjects in the images (people or dogs). After transformation, separate clusters seem to form for snow scenes involving a single person, groups of people, and dogs. Interestingly, images of dogs in fields and snow scenes are also drawn closer together. 
\textit{Softmax} (Figure~\ref{fig:tsne_softmax}) shows many small, isolated clusters before transformation. After transformation, bigger clusters seem to be created -- suggesting that the captions have again drawn related images together despite being different in the \textit{Softmax} space.
For \textit{bag of objects} (Figure~\ref{fig:tsne_boc}), objects seem to be clustered by co-occurrence of object categories, for example toilets and kitchens are clustered since they share sinks. Toilets and kitchens seem to be further apart in the transformed space.
We perform a similar analysis on the \emph{pseudorandom} representations (Figure~\ref{fig:tsne_pseudorandom}). We observe that the initial representations have very little explicit information and do not cluster well, indicating that the pseudorandom vectors are indeed noisy. The projected representations, however, form clusters that mimic the projected space of the BOO cluster, demonstrating that the model is able to factorize the noisy representations in the visual-semantic projection space guided by information from the captions. 
Enlarged versions of the images in Figure~\ref{fig:tsne} are also provided in the Appendix.%

\subsection{Domain dependency}
\label{sec:domaindependency}

We now demonstrate that end-to-end models are heavily reliant on datasets that have a similar training and test distribution. We posit that an IC system that performs similarity matching will not perform well on a slightly different domain for the same task. Demonstrating this will further validate our hypothesis that IC systems perform image matching to generate image captions.

We evaluate several models trained on MSCOCO on $1000$ test image samples from the \emph{Flickr30k}~\citep{young2014image} dataset~\footnote{the test split is obtained from http://staff.fnwi.uva.nl/d.elliott/wmt16/splits.zip}. Like MSCOCO, Flickr30k is an image description dataset; however, unlike MSCOCO, the images have a different object distributions and the captions are slightly longer and more descriptive. 

We evaluate the captions generated by our model with \emph{ResNet152} pool5 representation and by two other state-of-the-art models pretrained on MSCOCO: (a) Self-Critical (SC)~\citep{rennie2016self}, based on self critical sequence training that uses reinforcement learning, and (b) Bottom Up and Top Down (TDBU)~\citep{anderson2017bottom}, based on top-down and bottom-up attention using object region proposals. Both state-of-the-art models are much more complex than the image-conditioned RNN language model. The results are summarized in Table~\ref{tab:flickr}.

We observe that the scores drop by a large margin. A similar observation was made by~\citet{vinyals2016show}, and they alluded the drop in scores to the linguistic mismatch between the datasets. However, the out of training vocabulary words in the Flickr30k test set is only 8.6\%. This suggests that there is more to the issue than a mere vocabulary mismatch. Typical sentences on Flickr30k are structurally different and  generally longer, and the model is unable to generate good bigrams or even unigrams as is evident from B-1 and B-2 scores in Table~\ref{tab:flickr}. 

\section{Conclusions}
\label{sec:discussion}

We hypothesized that IC systems essentially exploit a \emph{distributional
similarity} space to `generate' image captions by attempting to match a test image to  similar training image(s) and generate an image caption from these similar images. Our study focused on the \emph{image} side of image captioning:
We varied the image representations while keeping the text generation component
of an end-to-end CNN-RNN model constant. We found that regardless of the image
representation, end-to-end IC systems seem to match images and generate
captions in a visual-semantic subspace for IC. We conclude that: %
\begin{enumerate}[(a)]
        \item End-to-end IC models are remarkably capable of separating structure
        from noisy input representations, as demonstrated by
        \textbf{pseudo-random vectors}; %
        \item  End-to-end IC models suffer virtually no significant loss in performance when a high dimensional representation is \textbf{factorized} to a lower dimensional space;
		\item End-to-end IC models can \textbf{learn a joint visual-textual semantic subspace} by clustering images with similar visual and linguistic information together; 
		\item End-to-end IC models rely on test sets having a \textbf{similar distribution} as the training set for generating good captions.
\end{enumerate}
The observations above strengthen our distributional similarity hypothesis -- that end-to-end IC models perform image matching and generate captions for a test image from similar image(s) from the training set -- rather than performing actual image understanding. Our findings provide novel insights into what end-to-end IC systems are actually able to do, which previous work only suggests or hints at without concretely demonstrating. We believe our findings are important for the community to further advance work on image captioning in a more informed manner. 

There is much scope for future work from the findings of this paper. One could examine the hidden states of the RNN model to better understand its behaviour and to further validate our distributional hypothesis. Understanding the theoretical formulation of the CNN-RNN architecture could also further help quantitatively confirm our hypothesis. Another useful direction would be to ascertain whether the distributional hypothesis also holds for more complex architectures, such as \cite{xu2015show,anderson2017bottom}; our intuition is that the hypothesis would remain valid even for such models. %

\paragraph{Acknowledgements.} This work is supported by the MultiMT project (H2020 ERC Starting Grant No. 678017). The authors also thank the anonymous reviewers for their valuable feedback on an earlier draft of the paper.

\bibliography{refs}
\appendix
\section{Hyperparameter Settings}
Our model settings were: 
\begin{itemize}
\item LSTM with 128 dimensional word embeddings and 256 dimensional hidden representations
Dropout over LSTM of 0.8
\item We used Adam for optimization. 
\item We fixed the learning rate to 4e-4
\end{itemize}

We report our results by keeping the above settings constant.

\section{Analyzing Transformed Image Representations: Enlarged Figures}

Figure~\ref{fig:tsne} shows an enlarged version of Figure 1 in the main paper for better viewing.

\begin{figure}[th]
  \begin{center}
    \begin{subfigure}[b]{0.80\linewidth}
      \begin{tabular}{c@{}c@{}c}
        \parbox[c]{0.45\linewidth}{\includegraphics[width=\linewidth]{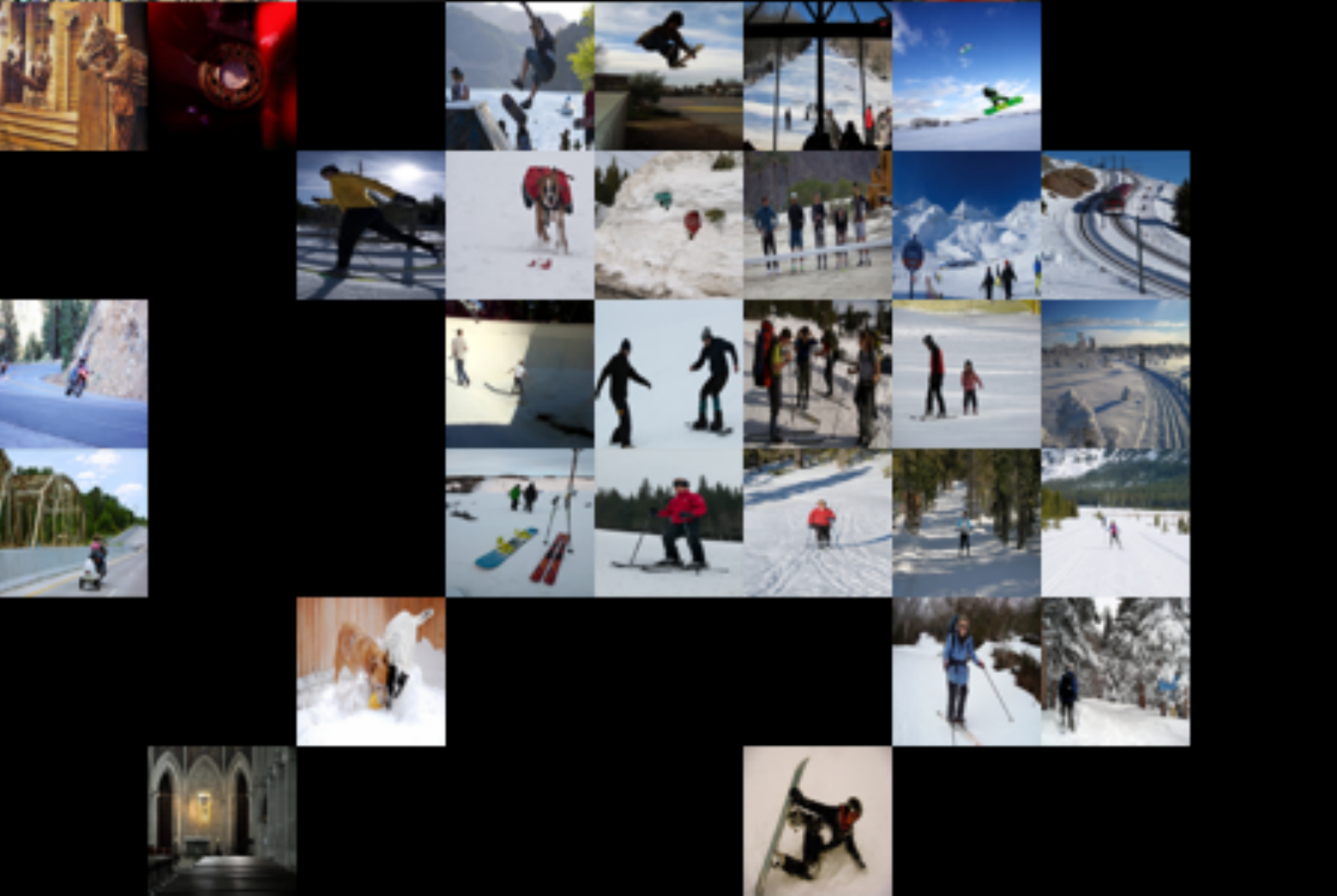}} &
    $\rightarrow$ &
        \parbox[c]{0.45\linewidth}{\includegraphics[width=\linewidth]{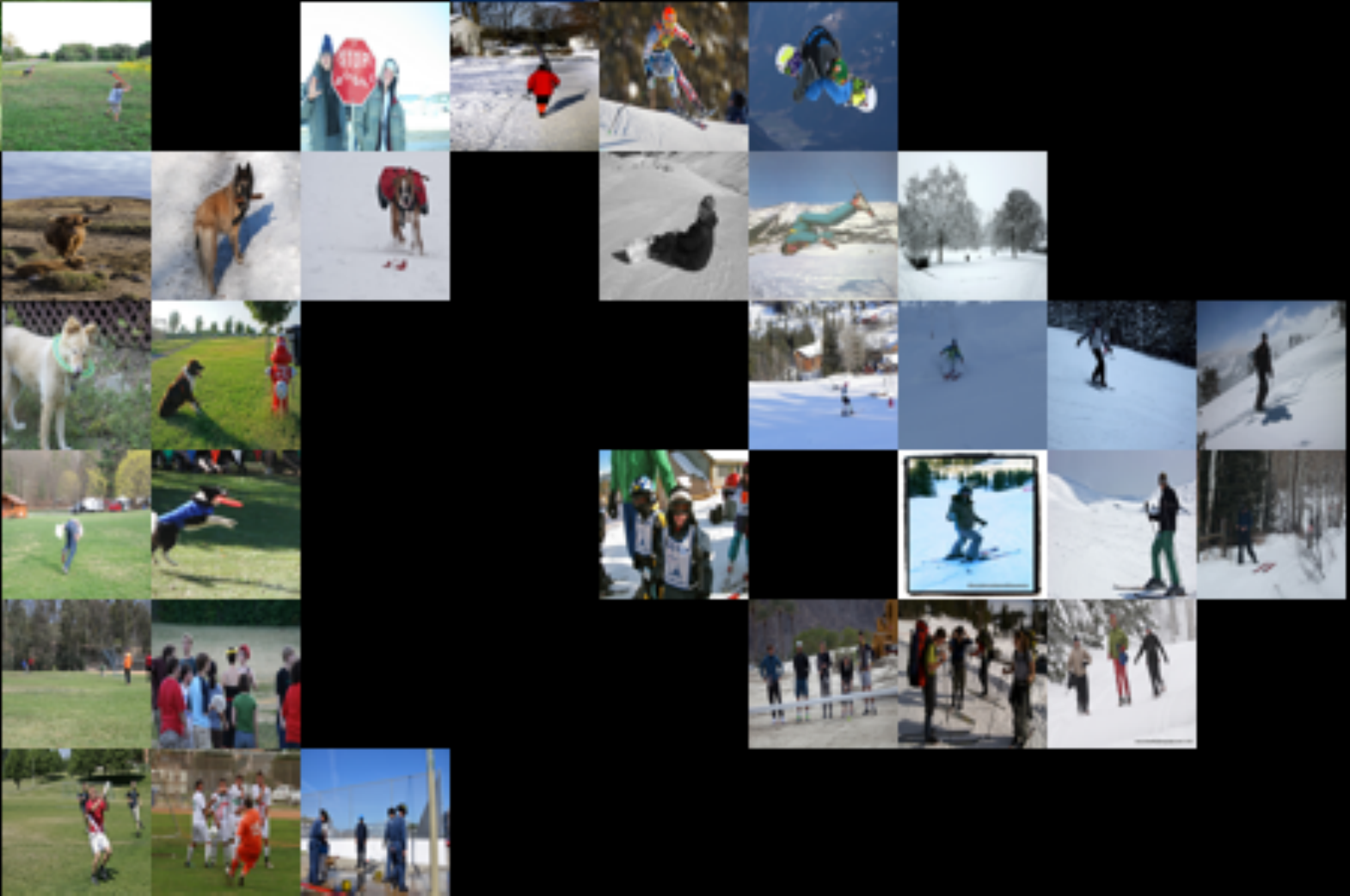}}
      \end{tabular}
      \caption{Pool5\label{fig:tsne_pool5}}
    \end{subfigure}
    \hfill
    \begin{subfigure}[b]{0.90\linewidth}
      \begin{tabular}{c@{}c@{}c}
        \parbox[c]{0.45\linewidth}{\includegraphics[width=\linewidth]{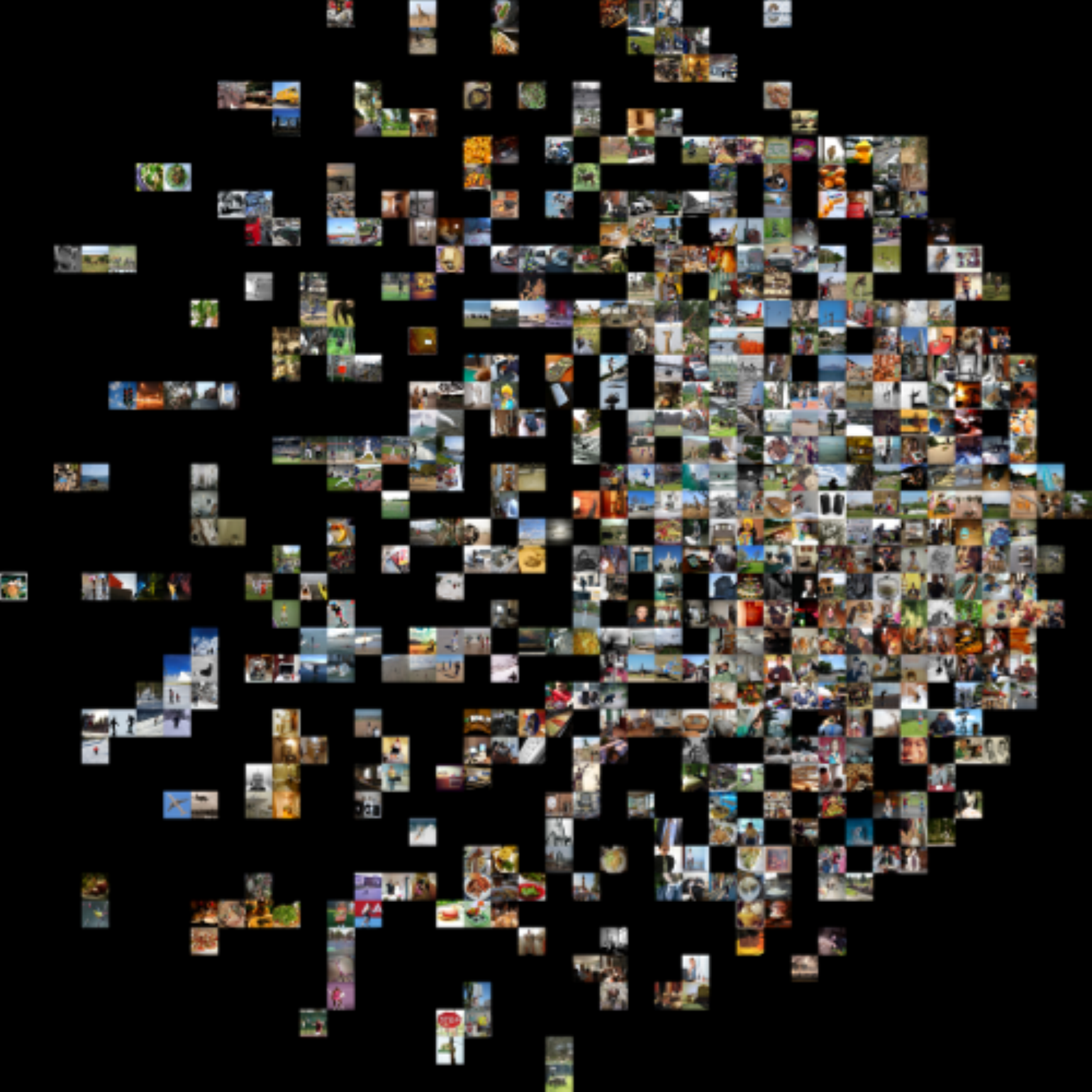}} &
    $\rightarrow$ &
        \parbox[c]{0.45\linewidth}{\includegraphics[width=\linewidth]{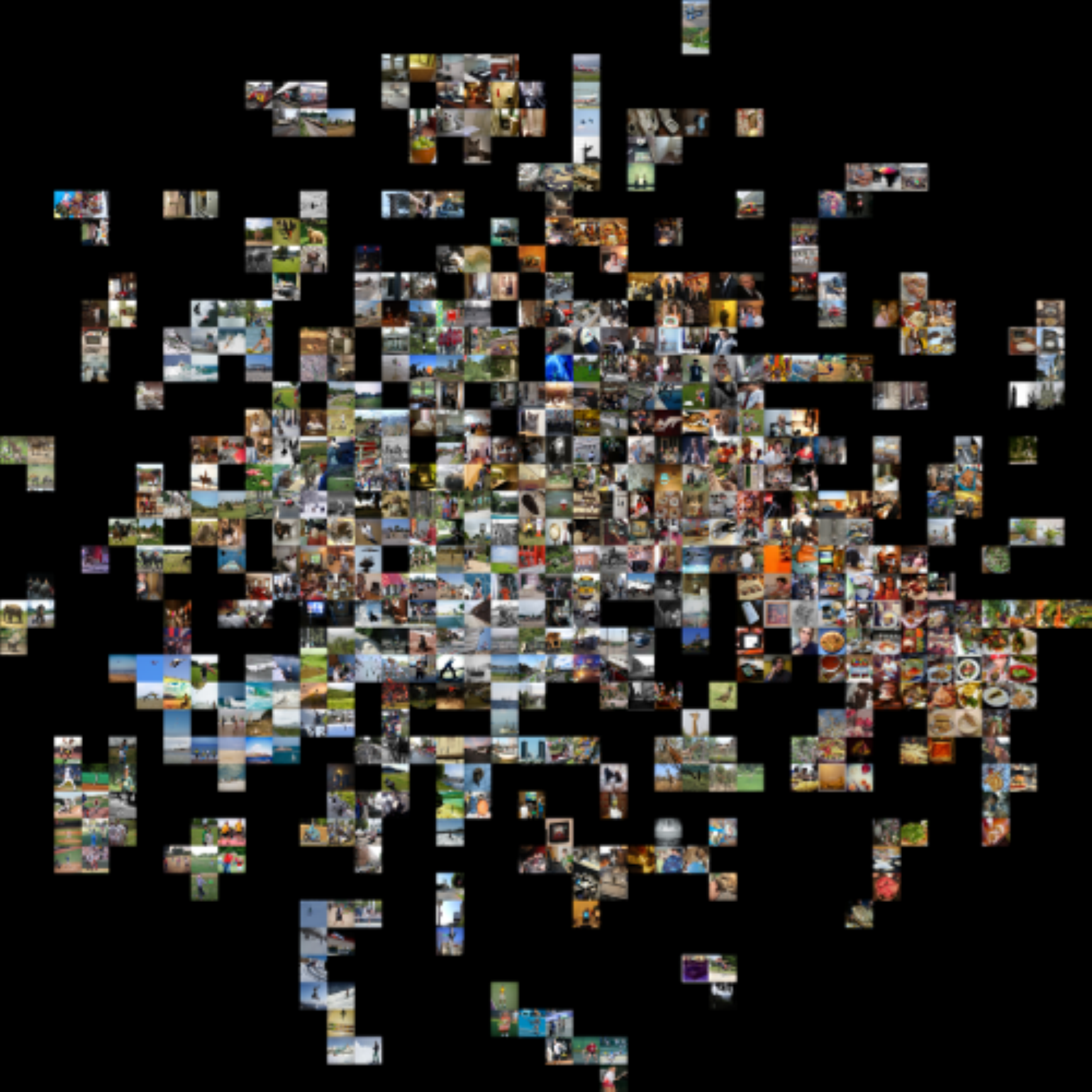}}
      \end{tabular}
      \caption{Softmax\label{fig:tsne_softmax}}   
    \end{subfigure}
    \begin{subfigure}[b]{0.80\linewidth}
       \begin{tabular}{c@{}c@{}c@{}c@{}c}
        \parbox[c]{0.30\linewidth}{\includegraphics[width=\linewidth]{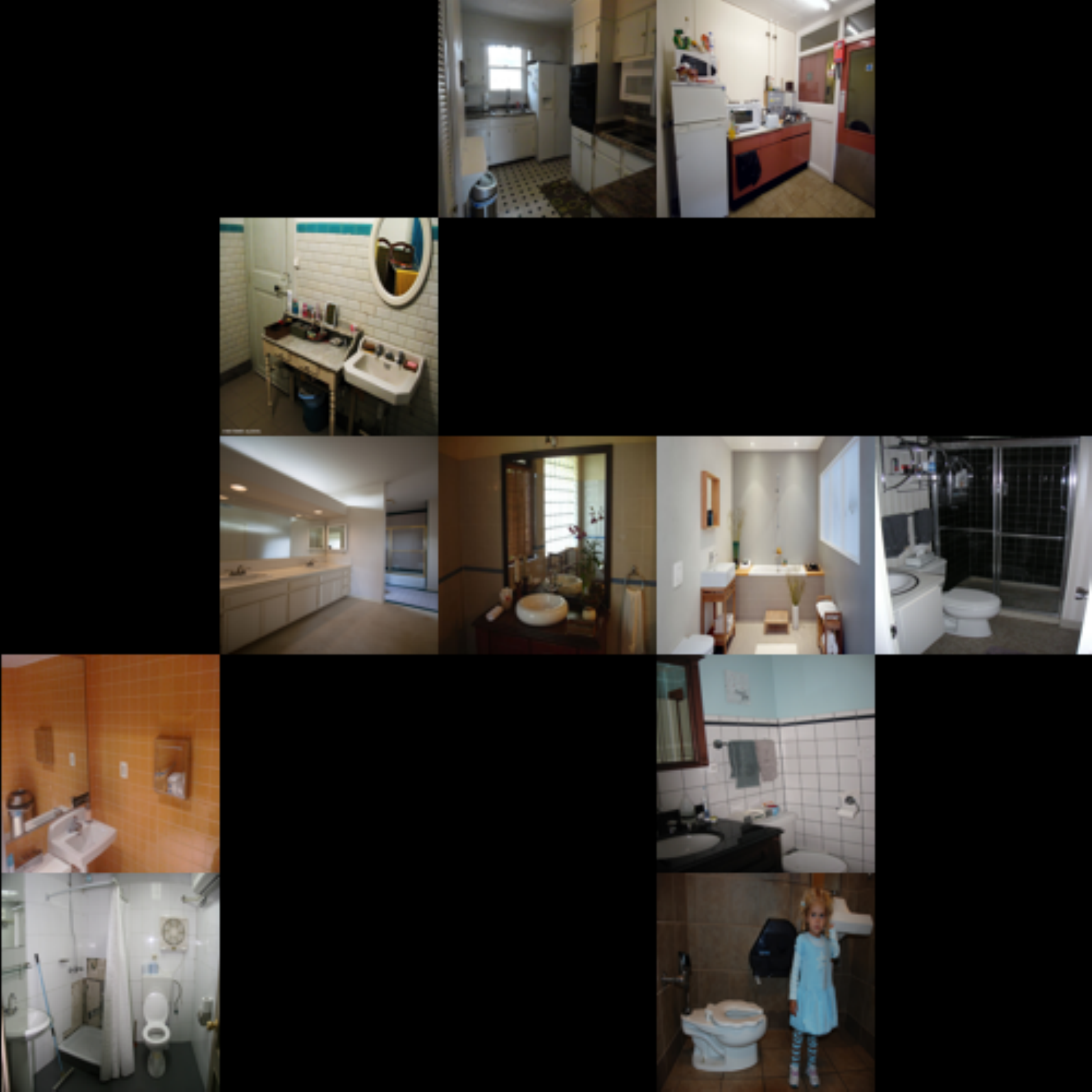}} &
         $\rightarrow$ &
        \parbox[c]{0.30\linewidth}{\includegraphics[width=\linewidth]{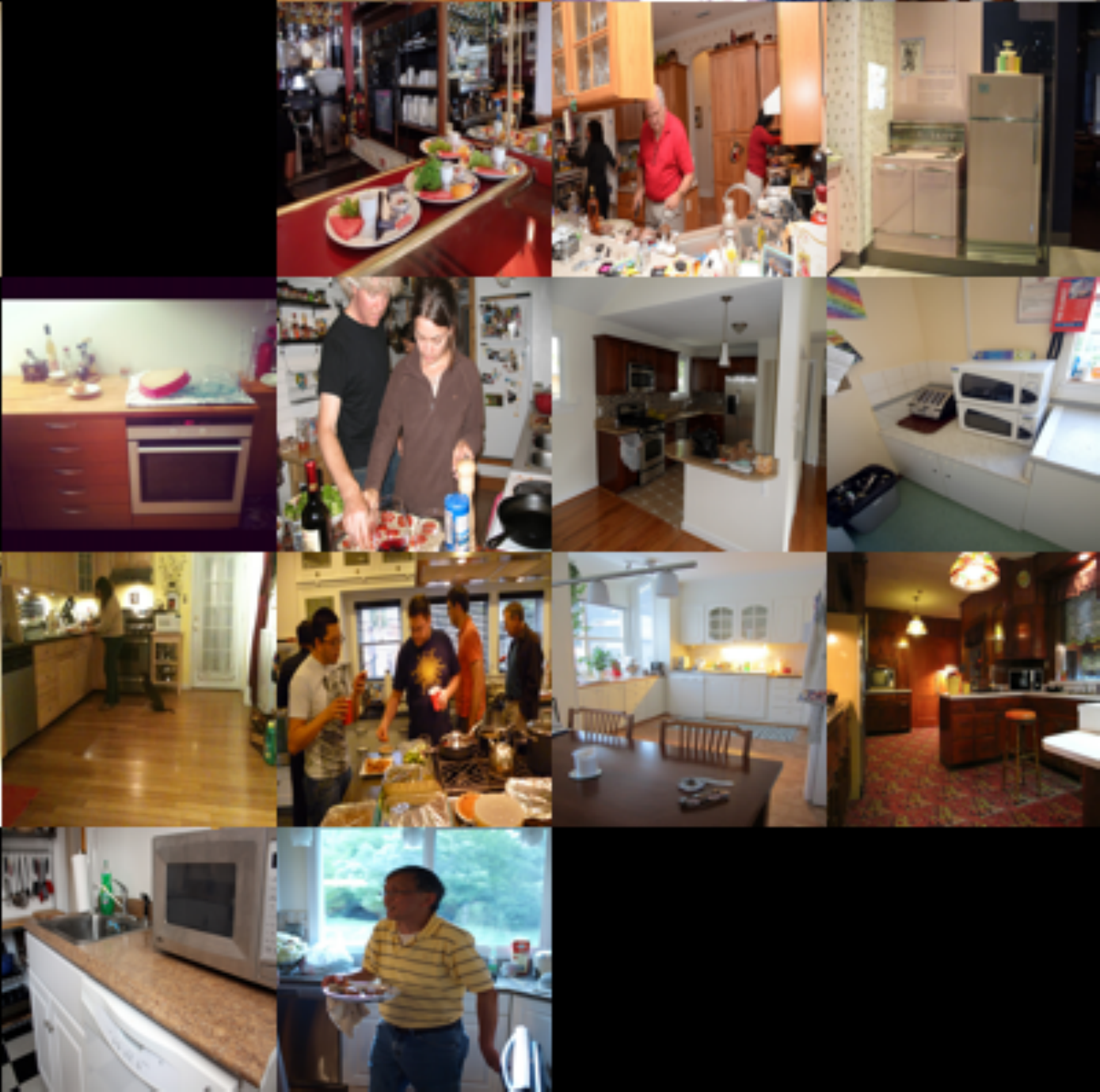}} &  
        ... &
        \parbox[c]{0.30\linewidth}{\includegraphics[width=\linewidth]{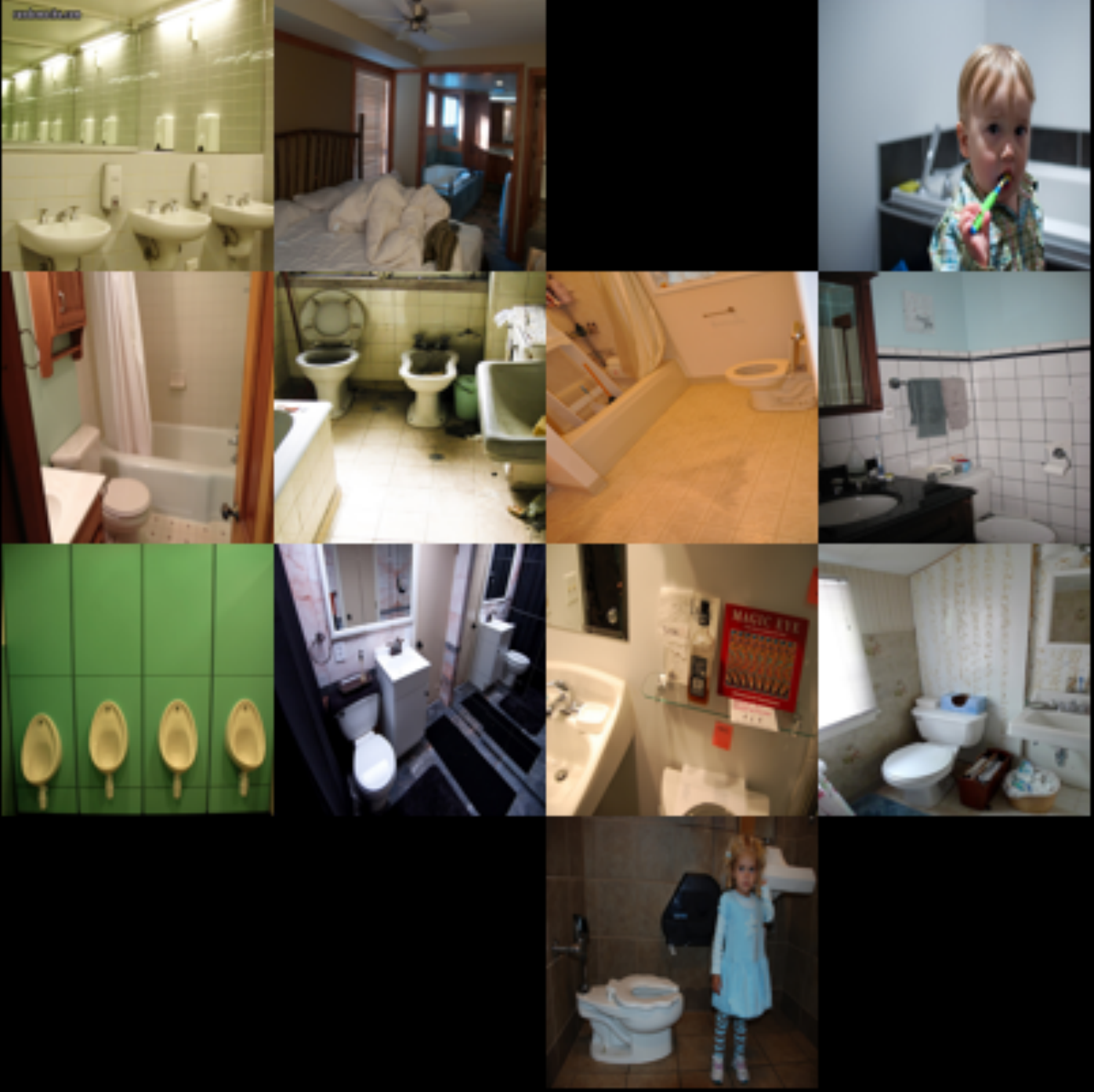}}     
      \end{tabular}
      \caption{Bag of objects\label{fig:tsne_boc}}   
    \end{subfigure}
    \begin{subfigure}[b]{0.70\linewidth}
       \begin{tabular}{c@{}c@{}c}
        \parbox[c]{0.45\linewidth}{
        \includegraphics[width=\linewidth]{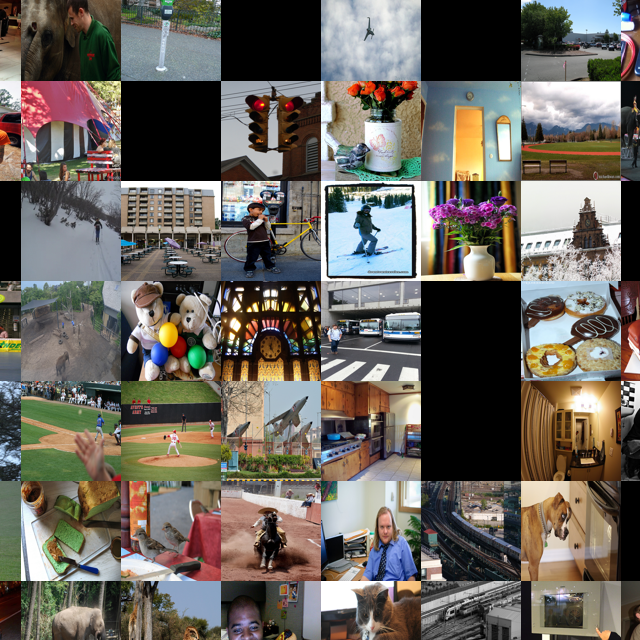}} &
    $\rightarrow$ &
        \parbox[c]{0.45\linewidth}{    
        \includegraphics[width=\linewidth]{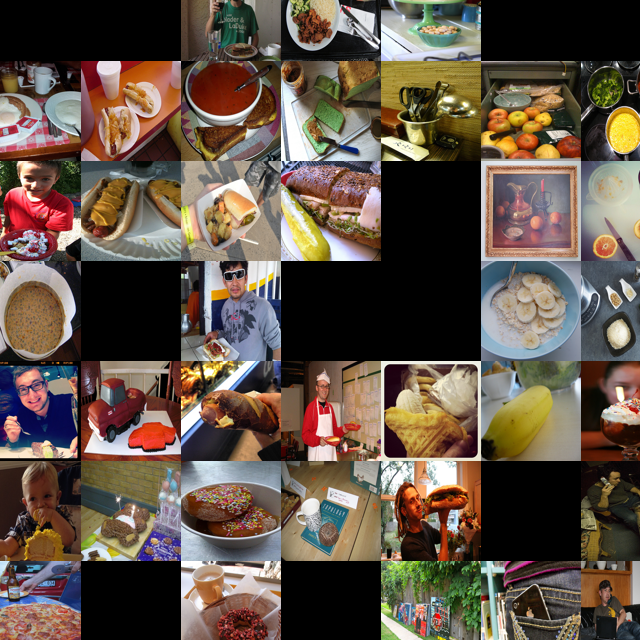}}
      \end{tabular}
      \caption{Pseudo-random\label{fig:tsne_pseudorandom}}
    \end{subfigure}    
  \end{center}
\caption{Visualization of the t-SNE projection of initial representational space (left) vs. the transformed representational space (right). See main text for a more detailed discussion. Please find the original images here: \url{https://github.com/sheffieldnlp/whatIC}}
\label{fig:tsne}
\end{figure}

\end{document}